\documentclass{article}

\usepackage{arxiv}

\usepackage[utf8]{inputenc} 
\usepackage[T1]{fontenc}    
\usepackage{hyperref}       
\usepackage{url}            
\usepackage{booktabs}       
\usepackage{amsfonts}       
\usepackage{nicefrac}       
\usepackage{microtype}      
\usepackage{lipsum}		
\usepackage{graphicx}
\usepackage{natbib}
\usepackage{doi}

\usepackage{color}
\usepackage{caption}
\usepackage{pifont}
\usepackage{dsfont}
\usepackage{footnote}
\usepackage{tabularx}
\usepackage{threeparttable}
\usepackage{amsmath,amssymb,amsfonts}
\usepackage{amsmath}
\usepackage{multirow}

\usepackage{caption}
\usepackage{subcaption}

\title{Human Activity Recognition from Wi-Fi CSI Data Using Principal Component-Based Wavelet CNN}

\author{ \href{https://orcid.org/0000-0000-0000-0000}{\includegraphics[scale=0.06]{orcid.pdf}\hspace{1mm}David S.~Hippocampus}\thanks{Use footnote for providing further
		information about author (webpage, alternative
		address)---\emph{not} for acknowledging funding agencies.} \\
	Department of Computer Science\\
	Cranberry-Lemon University\\
	Pittsburgh, PA 15213 \\
	\texttt{hippo@cs.cranberry-lemon.edu} \\
	\And
	\href{https://orcid.org/0000-0000-0000-0000}{\includegraphics[scale=0.06]{orcid.pdf}\hspace{1mm}Elias D.~Striatum} \\
	Department of Electrical Engineering\\
	Mount-Sheikh University\\
	Santa Narimana, Levand \\
	\texttt{stariate@ee.mount-sheikh.edu} \\
}

\author{Ishtiaque Ahmed Showmik\\
1606035@eee.buet.ac.bd
\And
Tahsina Farah Sanam\\
tahsina@iat.buet.ac.bd
\And
Hafiz Imtiaz\\
hafizimtiaz@eee.buet.ac.bd
}

\renewcommand{\shorttitle}{Preprint submitted to Elsevier}

\begin{document}
\maketitle
\begin{abstract}
Human Activity Recognition (HAR) is an emerging technology with several applications in surveillance, security, and healthcare sectors. Noninvasive HAR systems based on Wi-Fi Channel State Information (CSI) signals can be developed leveraging the quick growth of ubiquitous Wi-Fi technologies, and the correlation between CSI dynamics and body motions. In this paper, we propose Principal Component-based Wavelet Convolutional Neural Network (or PCWCNN) -- a novel approach that offers robustness and efficiency for practical real-time applications. Our proposed method incorporates two efficient preprocessing algorithms -- the Principal Component Analysis (PCA) and the Discrete Wavelet Transform (DWT). We employ an adaptive activity segmentation algorithm that is accurate and computationally light. Additionally, we used the Wavelet CNN for classification, which is a deep convolutional network analogous to the well-studied ResNet and DenseNet networks. We empirically show that our proposed PCWCNN model performs very well on a real dataset, outperforming existing approaches.
\end{abstract}

\section{Introduction}
Human activity recognition (HAR) is becoming extensively popular due to its abundant and far-reaching applications in smart homes, monitoring, and surveillance. HAR offers valuable insights into a person's physical functioning and behavior, which can be automatically monitored to provide individualized assistance. With the advancement of Wi-Fi technologies, individuals are now surrounded by devices capable of sensing and communication, which makes activity identification significantly efficient than image/video-based and wearable sensors-based approaches. More specifically, there are several techniques to recognize human activities: one strategy employs camera-based techniques, which have the drawbacks of requiring line of sight and suitable lighting conditions. Another strategy is using wearable sensors, which are more accurate and straightforward, but inconvenient and costly \cite{wang2015understanding}. Both of these approaches have the issue of potentially breaching personal privacy.

The use of Channel State Information (CSI) from Wi-Fi signals to detect human activity has developed rapidly in recent years. CSI measurement is a fine-grained metric that captures a Wi-Fi channel's amplitude and phase variations at distinct subcarrier levels. CSI is more stable than Received Signal Strength Indicator (RSSI); therefore, it is often the preferable option for implementing HAR systems \cite{sanam2020multi}. CSI signals allow us to observe activity across walls and behind closed doors and detect basic motions or a sequence of gestures based on the properties of signal penetration \cite{adib2013see}. Due to the unique capacity to decrease multipath effects, CSI signals carry useful information for activity recognition. As such, modern Wi-Fi-based device-free HAR systems exploit correlations between CSI signal changes and body motions \cite{wang2018spatial}. However, only a limited number of Wi-Fi receiving devices, such as Wi-Fi 5300 NICs, offer access to the CSI data \cite{gu2015paws}. CSI tools \cite{halperin2011tool} have made it feasible to capture CSI data and explore the association between signals and human activities. 

\subsection{Related Works} 
In recent years, researchers have developed several methods for CSI-based activity recognition. Yang et al. proposed a CSI signal enhancement method and antenna selection-based framework for human activity identification \cite{yang2021framework}. Wang et al. proposed a channel selective activity recognition system (CSAR), where they showed how the channel selection and the Long Short-Term Memory network (LSTM) can work together for HAR \cite{wang2018channel}. Gao et al. used a deep neural network to learn optimized deep features from radio images \cite{gao2017csi}. Hsieh et al. proposed a simple deep neural network for HAR \cite{hsieh2021end}, that incorporates the Multi-Layer Perceptron (MLP) and one-dimensional CNN (1D-CNN). Dayal et al. proposed a method with upgraded PCA technique and a deep neural network \cite{dayal2015human}. Yang et al. designed an occupancy detection system based on the CSI to demonstrate its effectiveness on the Internet of Things (IoT) platform \cite{yang2018device}. Another method that utilized Recurrent Neural Networks (RNN) for activity recognition was proposed by Ding and Wang \cite{ding2019WiFi}. Modern generative adversarial networks (GAN) are also being used for HAR, see the work of Wang et al. \cite{wang2021multimodal}, for example, which handled the problem of non-uniformly distributed unlabeled data with infrequently performed actions. Aside from these, two noteworthy works are WiKey and WiGest. WiKey is a keystroke recognition system proposed by Ali et al. \cite{ali2015keystroke} and WiGest is a in-air hand gestures detection system proposed by Abdelnasser et al. \cite{abdelnasser2015wigest} that employs the user's mobile device.  

Despite these advances in CSI-based activity recognition, there is still room for improvement in the robustness and effectiveness of the multi-class classification regime for HAR systems. More specifically, existing HAR systems have limited practical real-time applicability as they are affected by the high dimensional data. Additionally, CSI-based methods have their own set of difficulties and limitations. In this paper, we propose a unique approach to integrate PCA-based dimensionality reduction, DWT-based feature extraction and a CNN-based multi-class classification scheme that is capable of identifying activities regardless of the physiological and behavioral aspects of the individual and environmental elements, and is suitable for real-time applications.

\subsection{Our Contributions}
\begin{itemize}
    \item In this study, the subcarrier fusion method applying Principal Component Analysis (PCA) and the approach employing no fusion are explored. 
    \item An adaptive activity segmentation algorithm is implemented, which minimizes segmentation error and adapts the segmentation window to the data; imposing less constraints on the preprocessing.
    \item A novel HAR framework is proposed that integrates PCA-based dimensionality reduction, DWT-based feature extraction and CNN-based classification
    \item The proposed PCWCNN framework is contrasted to the baseline algorithms and contemporary HAR systems. In order to determine the model's robustness and flexibility, its performance in varied indoor environments is investigated. 
\end{itemize}
\section{Preliminaries}
In this section, we describe the necessary preliminary concepts that are relevant for our proposed method.

\subsection{Multipath Propagation Model}
A radio signal transmitted from a stationary source will encounter random objects in an indoor environment, producing replicas of the transmitted signal. When a single pulse is transmitted over a multipath channel, the received signal appears as a pulse train, with each pulse corresponding to a line-of-sight component or a different multipath component associated with different scatterers \cite{goldsmith2005wireless}. Because of the surroundings, there is one primary path (Line-Of-Sight, LOS) and multiple reflected paths (Non-Line-Of-Sight, NLOS). These NLOS components are duplicates of the transmitted signal that have been reflected, diffracted, or scattered and can be reduced in power, delayed in time, and phase and frequency displaced from the LOS signal path.

The Friis free space propagation model~\cite{wang2016wifall} is used to model the path loss of line-of-sight (LOS) path in a free space environment, applicable in the transmitting antenna's far-field region, and is based on the inverse square law of distance. The model expresses the received power $P_{r}(d)$ of a receiver antenna in terms of the distance $d$ from a radiating transmitter antenna in the free space:
\begin{equation}
    P_{r}(d)=\frac{P_{t} G_{t} G_{r} \lambda^{2}}{(4 \pi)^{2} d^{2}},
\end{equation}
where $P_t$ is the transmitted power, $G_r$ denotes the receiving antenna gain, $G_t$ denotes the transmitter antenna gain, $\lambda$ denotes the wavelength in meters and $d$ denotes the distance from the transmitter to a receiver in meters \cite{wang2016wifall}.
\begin{figure}[!ht]
\centering
\includegraphics[width = 0.4\textwidth]{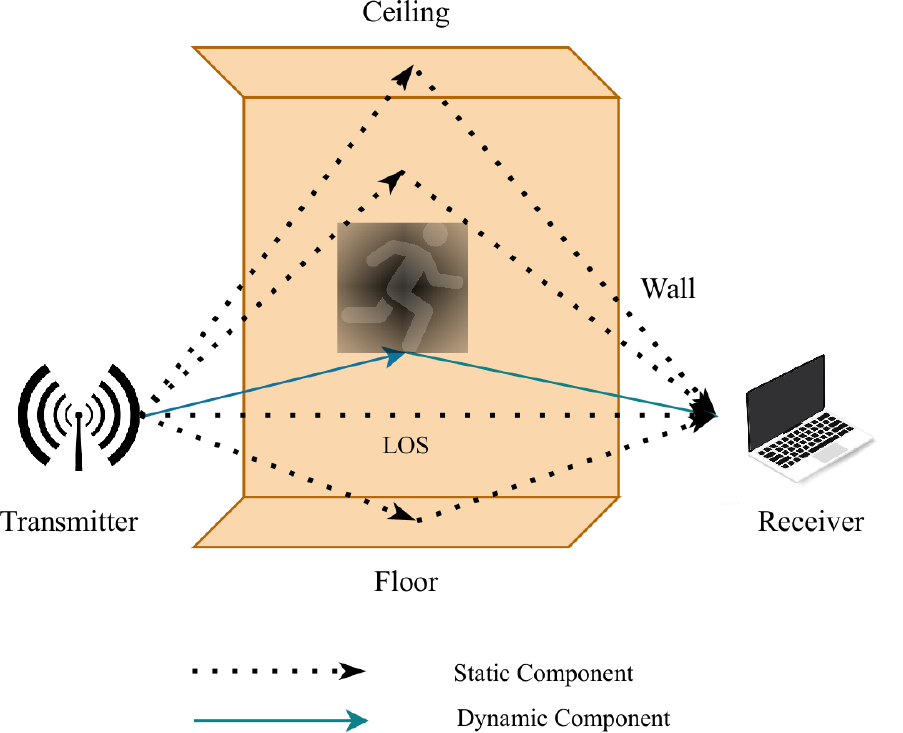}
\caption{Static and dynamic components of the multipath in an indoor environment caused by human activities}
\end{figure}

Let, $s(t)$ denote the transmitted signal as:
\begin{equation}
    s(t)=\Re\left\{a(t) e^{j\left(2 \pi f_{c} t+\phi_{0}\right)}\right\},
\end{equation}
where $a(t)$ is the time-varying complex-valued amplitude, $f_c$ is the carrier frequency. Then the total received signal is calculated as the sum of the LOS path and all the multipath components, and can be shown as:
\begin{equation}\label{multipath}
    r(t)=\Re\left\{\left[\sum_{n=0}^{N(t)} \alpha_{n}(t) e^{-j \phi_{n}(t)} a\left(t-\tau_{n}(t)\right)\right] e^{j 2 \pi f_{c} t}\right\},
\end{equation}
where $n = 0$ represents the LOS path. The number of resolvable multipath components, the multipath delay, the amplitude, and the phase shift due to the $n^{th}$ multipath are denoted by $N(t)$,  $\tau_n(t)$, $\alpha_n(t)$, and $\phi_{n}(t)$, respectively. Let's consider an indoor stationary environment where the objects are fixed. The expected values of the variables in the equation \eqref{multipath} will be time-invariant, assuming $a(t)$ is also time-invariant, as:
\begin{equation}
    A(t)=\sum_{n=0}^{N(t)} \alpha_{n}(t) e^{-j \phi_{n}(t)} a\left(t-\tau_{n}(t)\right).
\end{equation}
If an object is in motion in the environment, the multipath reflected or scattered from that object will have time-varying amplitudes. Let the multipath components corresponding to $n \in S$ have time-varying amplitudes. It can be shown that $A(t)=A_S+A_D(t)$, the received signal amplitude will be the sum of two components: the time-invariant static component, $A_S$, and the time-varying dynamic component $A_D$. The dynamic component will capture any motion occurring in the indoor environment. Therefore, the received CSI data will have both a static and a dynamic component. By focusing on the dynamic components at various subcarrier or frequency levels, it is possible to determine the association between the variation in human activity speed and the CSI dynamics.
\subsection{Overview of MIMO-OFDM Channel}
\begin{figure*}[!ht]
\centering

\includegraphics[width=0.28\textwidth]{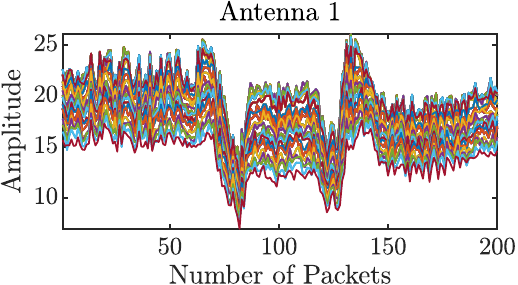}
\hspace{2em}
\includegraphics[width=0.28\textwidth]{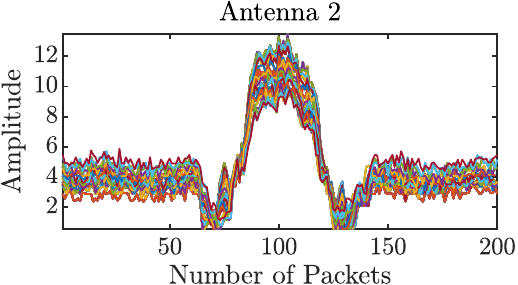}
\hspace{2em}
\includegraphics[width=0.28\textwidth]{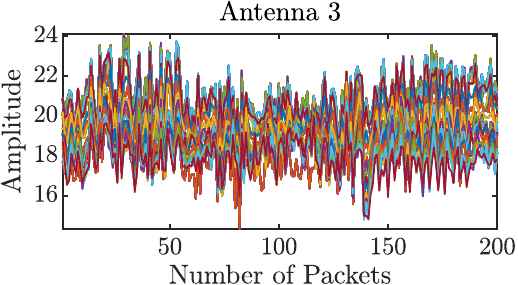}
\caption{CSI data of different Antennas}
\label{3ant3act}
\end{figure*}

\begin{figure*}[!ht]
\centering
\includegraphics[width=0.32\textwidth]{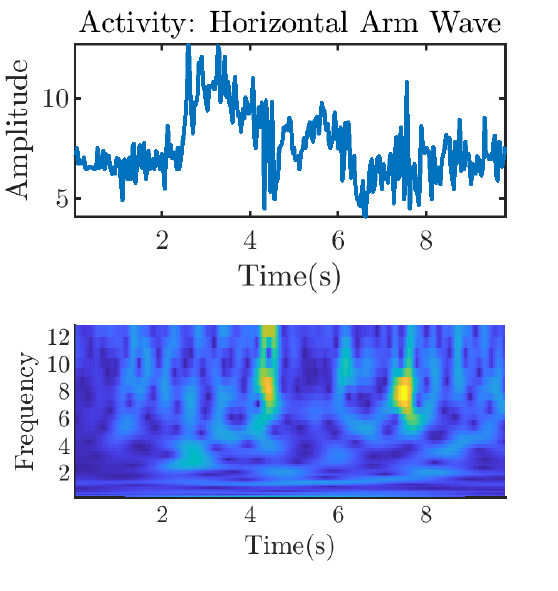}
\hfil
\includegraphics[width=0.32\textwidth]{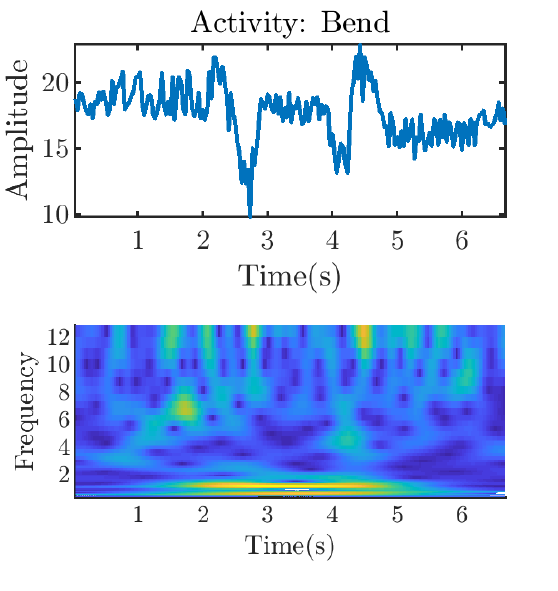}
\hfil
\includegraphics[width=0.32\textwidth]{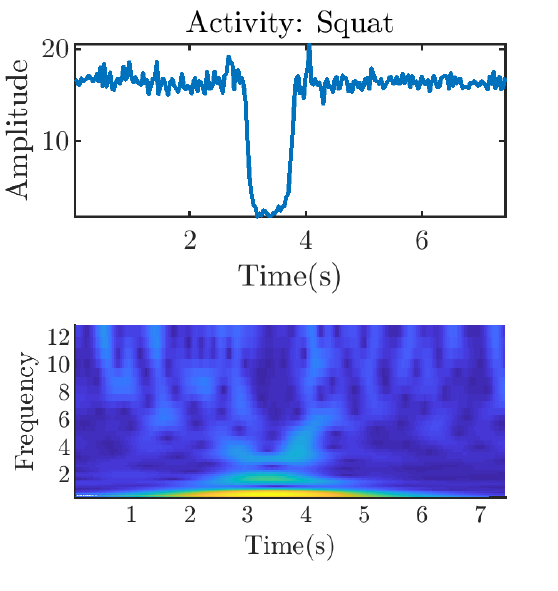}

\caption{CWT Scalogram of different activities}
\label{Time_Freq}
\end{figure*}
An array of several transmit or receive antennas is used in a Multiple Input Multiple Output (MIMO) communication system, where the array members are separated by distance. The same information is delivered across different fading channels. This is a strategy for increasing spatial diversity. If there are $N_T$ transmit and $N_R$ receive antennas, then $N_R \times N_T$ independent fading paths are available. The input-output relation of such a MIMO channel can be written as:
\begin{equation}
    \mathbf{y}=\mathbf{H s}+\mathbf{n},
\end{equation}
where $\mathbf{H}$ is an $N_R \times N_T$ channel matrix, $\mathbf{n}$ is the spatially and temporally white Zero-Mean Circulant Symmetric Complex Gaussian (ZMCSCG) noise vector with equal variance in each dimension, and $\mathbf{s}$ is the transmitted signal.

On the other hand, Orthogonal Frequency Division Multiplexing (OFDM) is a multicarrier modulation technology using densely spaced orthogonal subcarriers. Each of these subcarriers stands for a narrow-band flat fading channel. In the IEEE 802.11n standard, the CSI of an OFDM system provide amplitude and phase information for each subcarrier level \cite{sanam2020multi, sanam2019fuseloc}. 


\subsection{Channel State Information (CSI)}
Received Signal Strength Indicator (RSSI) and Channel State Information (CSI) are the two components of Wi-Fi signals. RSSI is the total energy of all signal channels combined, whereas CSI characterizes the properties of the wireless propagation channel between the transmitter and the receiver.
Because of multipath fading and background noise, RSSI measurements change over time and are therefore somewhat unreliable. On the other hand, CSI amplitude values exhibit significant consistency over time at a given point~\cite{sanam2020comute, sanam2018device}.
As mentioned before, the multipath propagation causes a delay spread in the time domain and selective fading in the frequency domain. These effects are incorporated in CSI, exhibiting channel selective fading on several subcarrier levels. The frequency response of the channel corresponding to each subcarrier can be expressed as:
\begin{equation}
    H(k)=|H(k)| e^{j \angle H(k)},
\end{equation}
where $H(k)$ denotes the CSI of $k^{th}$ subcarrier, $|H(k)|$
denotes the amplitude of the $k^{th}$ subcarrier, and $\angle H(k)$
denotes the phase.

CSI data can be extracted at the receiver as a $N_T \times N_R \times N_S$ matrix, where $N_T$ is the number of transmitting antennae, $N_R$ is the number of receiving antennae, and $N_S$ is the number of OFDM subcarriers. More specifically, CSI data can be expressed as:
\begin{align*}
CSI^1 &= [CSI^{1,1},\quad CSI^{1,2}, \ldots , CSI^{1,N_S}]          \\
CSI^2 &= [CSI^{2,1},\quad CSI^{2,2}, \ldots , CSI^{2,N_S}]          \\
\vdots \\
CSI^K &= [CSI^{K,1},\quad CSI^{K,2}, \ldots , CSI^{K,N_S}],          \\      
\end{align*}
where $K = N_RN_T$ is the number of streams present for each transmitter-receiver pair. Each pair contains a $N_S$ number of subcarriers, so the total number of CSI streams is $N_TN_RN_S$. Fig. \ref{3ant3act} shows the raw CSI data for a particular activity -- bending. The activity data consist of three-receiver antenna data, each of which further consists of 30 subcarrier data (20 subcarrier data has been plotted for simplicity). We observed that some antennas are more insensitive to some activity and as such, the CSI response of an insensitive antenna is not quite useful for activity recognition. Since the wireless channel is a frequency selective fading channel, different frequencies show different responses to the same activity. Therefore, CSI for each subcarrier is different. In general, subcarriers of low frequencies are less responsive to the activity, whereas subcarriers of high frequencies are more responsive \cite{sanam2018improved}. Moreover, low-frequency subcarriers are more prone to noise compared to high-frequency ones. If we arbitrarily choose a subcarrier for the purpose of HAR, the recognition performance may not be good. Therefore, further analysis is necessary for merging these subcarriers to remove redundancy, reduce noise and preserve maximum activity information. 
In an ideal situation, if a particular activity is considered for the same antenna and subcarrier, Wi-Fi CSI data should not differ much for different volunteers. But in practice, CSI data depends on many factors relating to the subject, namely the body shape, height, duration of activity performed, etc. Also, each individual's time to complete a particular activity is different. 



The objective of human activity recognition system modeling is to generate a robust system resilient to changes to the system, i.e., environmental changes, different subjects, etc. In an indoor environment, the Channel Frequency Response (CFR) is a linear combination of multipath components reflected off objects in the environment, including the subject's body \cite{wang2015understanding}. 
Thus, CSI data amplitude variation and phase change significantly depend on the surrounding environment. Therefore, it is crucial to distinguish the change in speed of the multipath components of each activity irrespective of the background and subject. Environmental changes significantly dominate statistical features. Time-Frequency analysis tools, such as Short-Time Fourier Transform (STFT) or Discrete Wavelet Transform (DWT), relate to the speed of the multipath changes \cite{wang2015understanding}.
Fig. \ref{Time_Freq} shows Continuous Wavelet Transform (CWT) scalogram for three activities. CWT has been chosen since it provides high-resolution analysis capability. These CWT scalograms can be considered a fingerprint of the activity being performed. In the scalogram of horizontal arm wave, there are high-frequency components of around 8 Hz and 12 Hz in 4-8 s time slot. For bend activity, significant low-frequency components in 2-5 s indicate the slow torso movement. Since upper body movement is also present, there are few high-frequency components. On the other hand, for the last activity, squat, there are only low-frequency components in 2-5 s, indicating slow lower body movement. The sampling rate of WiAR CSI data  was 30 Hz. So the highest frequency to show on the scalogram is Nyquist frequency 15 Hz.

The primary challenge with CSI-based activity recognition is
that CSI data contains noise, making it difficult to use any
detection system directly on the raw data. To eliminate these
artifacts, several pre-processing strategies are required. The
second challenge is choosing a proper feature extraction approach.
For the same activity, different individuals need varying times-
pans. The duration of an activity might change over time, even
for the same individual. Moreover, a segmentation task is required to identify the activity part; however, inaccurate segmentation
may lead to low accuracy. This study proposes a technique
that addresses these issues and yields acceptable outcomes. 

\subsection{Residual Networks and Dense Convolutional Networks}
Resnets \cite{he2016deep} provided a feedforward neural network implementation with "shortcut connections" or "skip connections," bypassing one or more layers when deeper networks may start converging. This approach addresses the issue of loss or saturation of accuracy as network depth increases. Deeper networks, even with shortcut connections, face the difficulty of quickly disappearing information about the input and gradient as it propagates through the networks \cite{fujieda2018wavelet}. To overcome this issue, Dense Convolutional Network (DenseNet) \cite{huang2017densely} incorporates shortcut connections connecting each layer to its preceding layers. Through channel-wise concatenation, dense connections allow each level of deconstructed signal to be directly coupled with all following levels. With this connection, the network can effectively route all information from the input side to the output side. 

\section{System Overview}
\begin{figure*}[!ht]
\centering
\includegraphics[width = 0.95\textwidth]{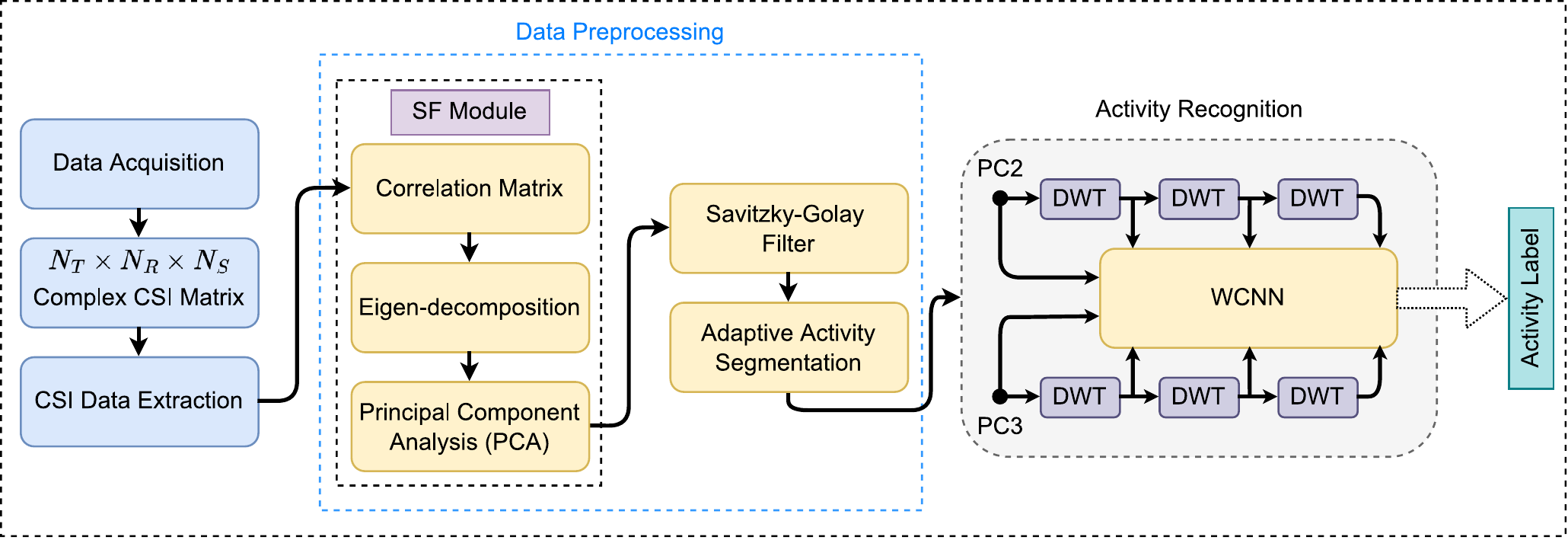}
\caption{Framework of PCWCNN activity recognition system}
\label{framework}
\end{figure*}
Our proposed PCWCNN system framework is presented in Fig. \ref{framework}. The modules that are included in this system are the CSI Data Extraction module, Subcarrier Fusion (SF) module, Savitzky-Golay Filter, Adaptive Activity Segmentation module, and PCWCNN model. We describe each of these components below.

\subsection{Subcarrier Fusion (SF) Module}
The authors in \cite{wang2015understanding} developed a way of applying PCA to CSI streams to solve the problem of integrating CSI streams and reducing noise since typical frequency domain filters are ineffective. The steps involving this method are finding the correlations between CSI streams, performing an Eigen decomposition, and determining the principal components.

Let $\mathbf{X_i}$ denote the $i^{th}$ CSI stream ($i^{th}$ subcarrier of the CSI data. We can arrange all the CSI streams column-wise to form the matrix as
$
\mathbf{X} = [\mathbf{X_{1}} , \mathbf{X_{2}} , \mathbf{X_{3}} , \cdots , \mathbf{X_{N}}].
$
We can find the correlation estimate from the auto-correlation matrix
$
\mathbf{R} = \mathbf{X}^{T}\mathbf{X} \in \mathbb{R}^{N\times N}.
$
As the Eigenvectors of an auto-correlation matrix are orthogonal to each other, it is possible to project the data on the eigenvectors to find the components in their signal subspace. The $k^{th}$ principal component can be constructed as
$
\mathbf{h}_{k} = \mathbf{R}\mathbf{v}_{k}.
$ 
If we arrange the eigenvectors according to the decreasing order of their corresponding eigenvalues, then the smallest few eigenvalues will be associated with the noise and the principal components corresponding to these eigenvalues can be ignored. This will effectively filter out a portion of the noise. Discarding these principal components will also result in dimensionality reduction. Additionally, the first principal component has the largest possible variance. Although the first  principal component captures most of the data variability, it simultaneously captures the burst noise in all the CSI streams. As discussed in \cite{wang2015understanding}, noises caused by internal state changes are most prominently present in the first principal component. Therefore, we discard the first principal component and utilize the second and third principal components in our method. The selected principal components will be subjected to further processing and denoising using Savitzkey-Golay Filter.

\subsection{Savitzkey-Golay Filter}
Savitzky and Golay \cite{savitzky1964smoothing} presented a data smoothing method based on local least-squares polynomial approximation. They demonstrated that discrete convolution with a fixed impulse response is similar to fitting a polynomial to a set of input samples and then evaluating the resultant polynomial at a single point inside the approximation interval. The low pass filters obtained by this method are known as Savitzky-Golay filters \cite{schafer2011savitzky}. 

A set of $2M+1$ input samples within the approximation interval are effectively combined by a fixed set of weighting coefficients that can be computed once for a given polynomial order $N$ and approximation interval with a length of $2M+1$. If $\tilde{p}[n]$ is the polynomial fit to the unit impulse evaluated at the integers $-M\le n\le M$, then
\begin{equation}
\tilde{p}[n] = \sum_{k = 0}^{N}\tilde{a_{k}}n^{k},
\end{equation}
where,
\begin{equation}\tilde{\mathbf{a}} = \left( \mathbf{A}^{T} \mathbf{A}\right)^{-1}\mathbf{A}^{T}\mathbf{d},\end{equation}, $\mathbf{d} = [0, 0, ..., 0, 1, 0, ..., 0, 0]^{T}$ is a $(2M+1)\times1$ column vector impulse and $\mathbf{A}^{T}$ is the $(N+1)\times(2M+1)$ matrix:
\begin{equation}
\mathbf{A}^{T} = 
\begin{bmatrix}
\left( -M \right)^{0} & \cdots  & (-1)^{0} & 1 & 1^{0} & \cdots  & M^{0} \\
\left( -M \right)^{1} & \cdots  & (-1)^{1} & 0 & 1^{1} & \cdots  & M^{1}\\
\left( -M \right)^{2} & \cdots  & (-1)^{2} & 0 & 1^{2} & \cdots  & M^{2}\\
\vdots                & \vdots  & \vdots &  \vdots& \vdots  & \cdots  & \vdots \\
\left( -M \right)^{N} & \cdots  & (-1)^{N} & 0 & 1^{N} & \cdots  & M^{N}.
\end{bmatrix}
\end{equation}
The impulse response of the filter is:
\begin{equation}
h[-n] = \tilde{p}[n].
\end{equation}

\begin{figure*}[!ht]
\centering
\includegraphics[width = 0.88\textwidth]{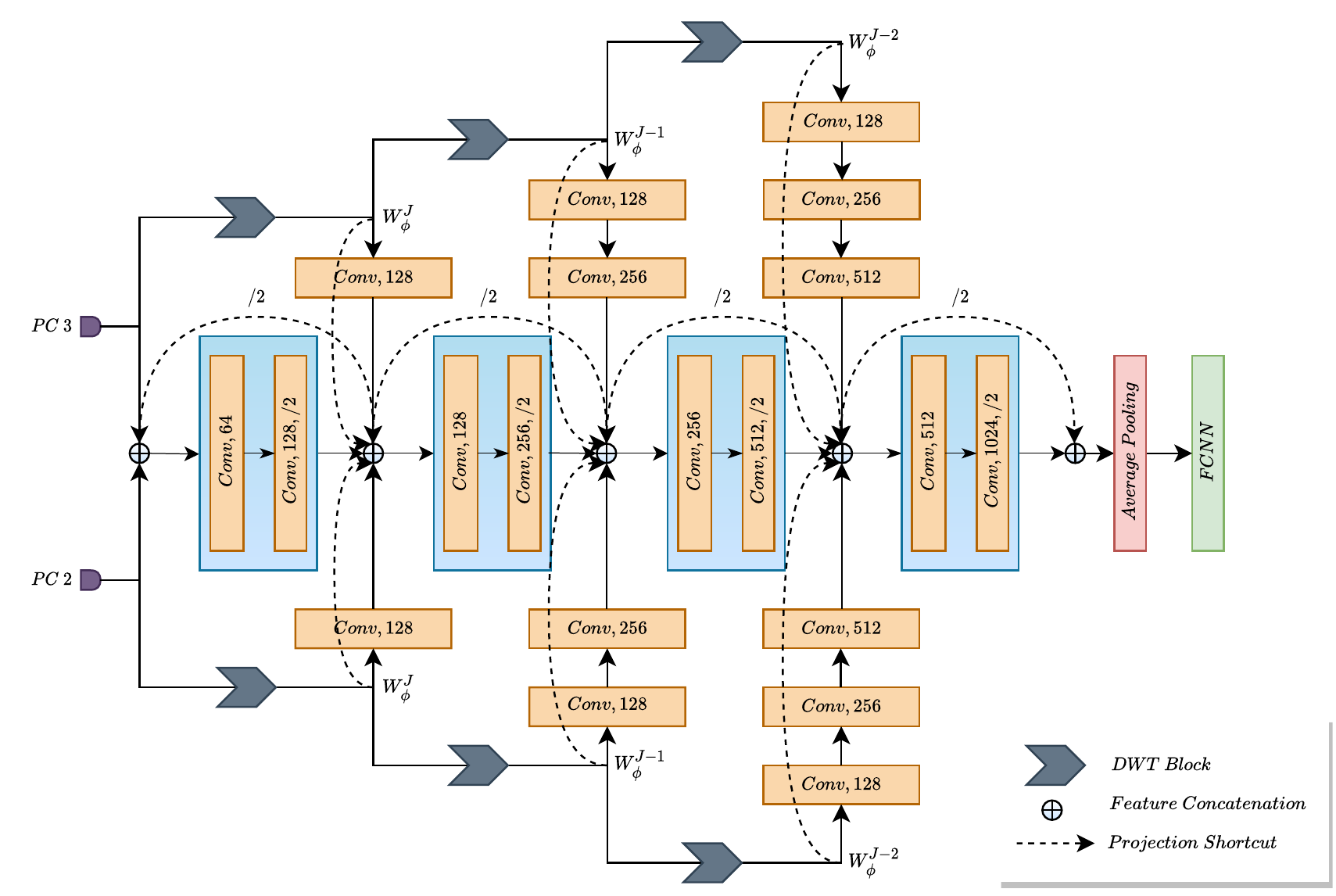}
\caption{Activity detection network of the proposed PCWCNN framework: This network provides a three-level multi-resolution decomposition of both principal components. Deep CNN is used, which consists of convolution layers with $3\times1$ kernels. To minimize feature map dimensions, $3\times1$ convolutional kernels, a stride of $2$ and $1\times1$ padding are employed. The level $J$ approximation coefficient $W_{\phi}^J$  and the convolutional layers are channel-wise concatenated at each stage. Conv is followed by a number indicating the number of output channels. Global average pooling followed by flattening is implemented, which is then followed by a fully connected layer (FCNN). The last layer is the softmax layer, which generates activity class estimates.}
\label{wcnn_model}
\end{figure*}

\subsection{Adaptive Activity Segmentation}
The precise segmentation of the activity part from CSI data is a challenging aspect of activity recognition. Activity segmentation is necessary to distinguish the dynamic activity segment from the static non-activity section. Non-activity data is excluded from the recognition process for improved efficiency since it contains no information about the activity. To that end, a sliding window-based segmentation algorithm is proposed in \cite{yang2021framework}. It finds an array of the mean-variance of raw CSI data and filters out the part lower than a threshold corresponding to the third quartile of the sorted variance values. There are certain limitations to such an approach. Some activities may be erroneously segmented with a non-adaptive window since they are performed for a different duration. Some activities may last longer than the segmented window, whereas others may be of less duration. As a result, an adaptive algorithm is required to modify the threshold value according to the nature of the signal. A modified sliding window-based segmentation process has been used to solve this issue. After the sliding window is applied to the denoised principal components, the variance in that window is computed and saved in $\mathbf{V}$. Then, the sliding window is shifted across $\mathbf{V}$, and the mean in that window is calculated and stored in $\mathbf{M}$. The values of $\mathbf{M}$ are sorted and the maximum value is obtained. A threshold value $T = p \times \max{(\mathbf{M})}$ is computed where $0<p<1$, and the values in $\mathbf{M}$ less than $T$ are discarded. Value of $p$ is set such that the segmentation length $t$, normalized to the length of the CSI data sequence is greater than some prefixed value $t_1$ and less than $t_2$, and $0<t_1<t<t_2<1$. In this work, the value of the parameters $t_1$ and $t_2$ were selected to be 0.3 and 0.5, respectively. The value of normalized segment length $t$ is minimized in an iterative manner. The start time and end time of the activity segment are obtained from the filtered $\mathbf{M}$. Once the segmented activity parts of PC2 and PC3 have been obtained, activity recognition followed by feature extraction is performed.

\subsection{DWT-Based Feature Extraction}
Because the statistics of CSI signal frequency vary with time, CSI activity signal can be better represented on a time-frequency domain concurrently. To that end, we use the Discrete Wavelet Transform (DWT) in this work. The DWT, is a computationally efficient approach for extracting information from non-stationary signals. In contrast to the Short-time Fourier Transform (STFT), which provides uniform time resolution across all frequencies, DWT offers high time resolution and low-frequency resolution for high frequencies and high-frequency resolution and low time resolution for low frequencies \cite{tzanetakis2001audio}. The DWT coefficients of a discrete sequence $x(n)$ can be written as
\begin{align}
    & W_{\phi }(j_{0},k)  = \frac{1}{\sqrt{N}}\sum_n {x(n)\phi_{j_0,k}(n)}, \\
    & W_{\psi}(j,k)  = \frac{1}{\sqrt{N}}\sum_n {x(n)\psi_{j,k}(n)} \quad for \; j \geq j_0.
\end{align} 
 Here, ${\phi }(j_{0},k)$ is the scaling basis function and ${\psi }(j,k)$ is the wavelet basis function. Generally, $j_0$ is set to be 0 and N to be a power of 2, which is the length of the discrete sequence $x(n)$. The approximation coefficients are obtained by the convolution of the input signal $x(n)$ with the scaling filter, followed by dyadic decimation. Similarly, the detail coefficients are obtained by the convolution of the input signal $x(n)$ with the wavelet filter, followed by dyadic decimation:
 \begin{align}
 & W_{\psi}(j,k) = \sum_n {h_\psi(n-2k)W_{\phi}(j+1,n)}, \\
 & W_{\phi}(j,k) = \sum_n {h_\phi(n-2k)W_{\phi}(j+1,n)},
 \end{align}
 where $W_{\phi}(j,k)$ and $W_{\psi}(j,k)$ are the level $j$ approximation and detail coefficients, and $h_\phi(n)$ and $h_\psi(n)$ are the scaling and wavelet filters.

\subsection{Activity Recognition}
Fujieda et al. \cite{fujieda2018wavelet} proposed Wavelet Convolutional Neural Network (Wavelet CNN); they demonstrated that wavelet CNNs produce comparable, if not better, accuracies with a substantially fewer trainable parameter than conventional CNNs. Wavelet CNN, or WCNN, is thus less complicated to implement and is more suitable for real-time applications. Furthermore, it is less susceptible to over-fitting and consumes less memory than a typical CNN. WCNN is influenced by well-known architectures such as Residual Networks (ResNets) and Dense Convolutional Networks (DenseNets). 

\subsection{Summary of the Proposed System}
Following the PCA and organizing the Eigenvectors in decreasing order of their associated eigenvalues, the first principal component and those corresponding to lesser eigenvalues were removed in the SF module. Discarding these unwanted principal components and keeping only the second and third principal components, dimensionality reduction was achieved along with noise reduction. One activity sample from the dataset used in this study yielded a CSI with 90 subcarriers, of which only two principal components remained following subcarrier fusion. 

In this work, we propose a modified WCNN network, namely the Principal Component-based Wavelet Convolutional Neural Network (PCWCNN). This model takes the second and third principal components from the SF module and feeds them into the WCNN, and a three-level multi-resolution decomposition of both principal components is performed. The lower resolution counterpart of the DWT output captures the bulk of the original signal since the information in the fine-grained signal is typically sparse. As a result, the low-frequency version or approximation coefficients of the DWT output are used throughout this network.

The original principal components are concatenated and fed as input into a deep CNN network. The approximation coefficients from one step of the wavelet transformation are iteratively passed through multiple identical transformations to yield various frequency renditions of the original principal components. Using the feature concatenation method, the approximation coefficients of these stages are combined with the feature maps of the convolutional layers across the CNN. The CNN consists of convolution layers with $3\times1$ kernels. To minimize feature map dimensions, $3\times1$ convolutional kernels, a stride of $2$ and $1\times1$ padding are employed. Incorporating the ResNet and Densenet architectures, along with wavelet decomposition, may significantly improve the recognition accuracy of a deep CNN \cite{fujieda2018wavelet}. Because DWT inherently comprises a decimation of the input signal, the approximation coefficient size must be analogous to that of the feature maps of the convolutional layers across the CNN to employ the feature concatenation approach. A method can be used that employs $1\times1$ padding with a stride of 2 to halve the output to the size of the input layer \cite{fujieda2018wavelet}. Without reducing accuracy, this strategy may be used instead of max-pooling~\cite{springenberg2015striving, fujieda2018wavelet}.  We used global average pooling followed by flattening, which is then followed by a fully connected layer (FCNN). The softmax layer generates activity class estimates.

\section{Implementation and Evaluation}
This section describes the proposed model's implementation and evaluation compared to reference models. This study used the WiAR data set \cite{guo2019wiar} for activity recognition leveraging Wi-Fi CSI data. This is a publicly accessible data set that focuses on different types of activities and environments. WiAR gathers Wi-Fi CSI data from three locations. The performance of the recognition method is influenced by the environment in which it is implemented.\\
The WiAR dataset comprises sixteen activities that may be grouped into three major categories based on the body parts associated: upper body, lower body, and entire body activities. Upper body activities are those in which volunteers do the task primarily with their upper skeleton joints. Lower-body movements only move the lower skeleton joints, entire body activities combine upper and lower body activities. The data set is more diverse when there are multiple forms of categorization. WiAR dataset acquired more than 7s of data for each activity sample, which includes 2-3s of activity data and 4-6s of no-activity data (static component from the environment). In general, it takes a volunteer about 2-3 seconds to complete an activity at a moderate pace. 

\subsection{Data Acquisition}
\begin{figure*}[!ht]
         \centering
         \begin{subfigure}[b]{0.46\textwidth}
         \includegraphics[width = \textwidth]{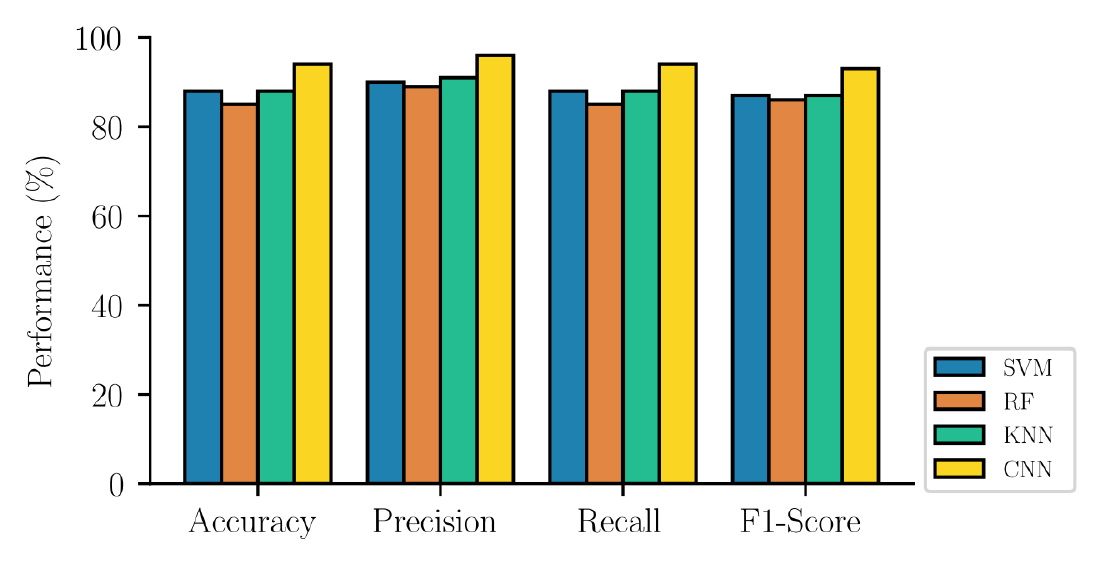}
         \caption{}
         \end{subfigure}
         \begin{subfigure}[b]{0.46\textwidth}
         \includegraphics[width = \textwidth]{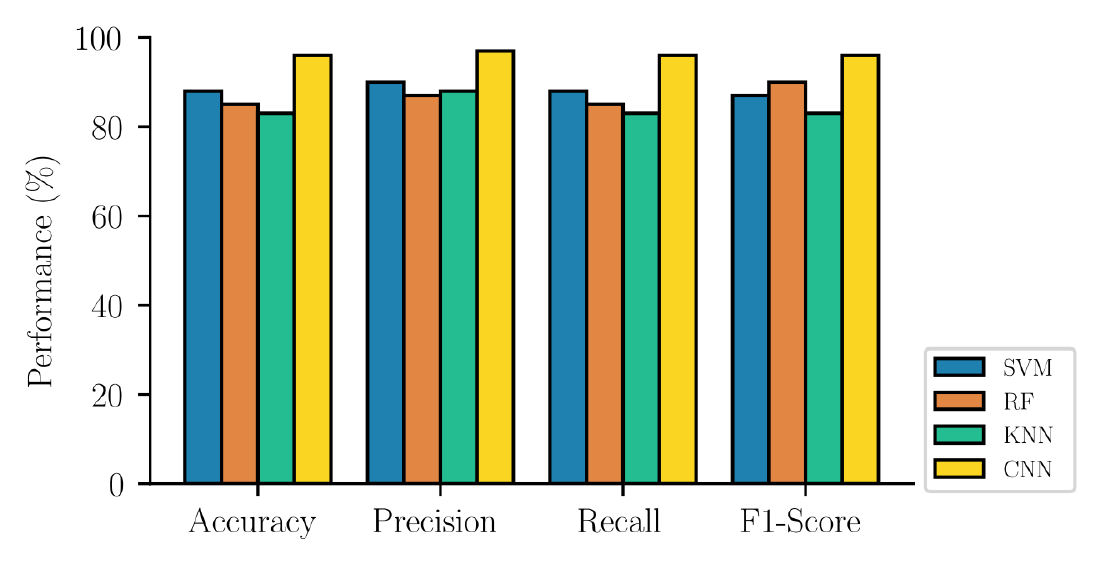}
         \caption{}
         \end{subfigure}

\caption{Performance comparison of different approaches on an individual employing (a) all subcarriers (b) subcarrier fusion}
\label{pca_dwt_fig}
\end{figure*}

\begin{table*}[!ht]
\caption{Precision, Recall and F1-Score comparison of different approaches on one volunteer employing SF  module and DWT feature extraction. }
\label{pca_dwt}
\resizebox{\textwidth}{!}{
\begin{tabular}{|c|ccc|ccc|ccc|ccc|}
\hline
\multirow{2}{*}{\begin{tabular}[c]{@{}c@{}}Activity\\ Class\end{tabular}} & \multicolumn{3}{c|}{SVM}                                                & \multicolumn{3}{c|}{RF}                                                 & \multicolumn{3}{c|}{KNN}                                                & \multicolumn{3}{c|}{CNN}                                                \\ \cline{2-13} 
                                                                          & \multicolumn{1}{c|}{Precision} & \multicolumn{1}{c|}{Recall} & F1-Score & \multicolumn{1}{c|}{Precision} & \multicolumn{1}{c|}{Recall} & F1-Score & \multicolumn{1}{c|}{Precision} & \multicolumn{1}{c|}{Recall} & F1-Score & \multicolumn{1}{c|}{Precision} & \multicolumn{1}{c|}{Recall} & F1-Score \\ \hline
01                                                                       & \multicolumn{1}{c|}{1.00}      & \multicolumn{1}{c|}{1.00}   & 1.00     & \multicolumn{1}{c|}{0.75}      & \multicolumn{1}{c|}{1.00}   & 0.86     & \multicolumn{1}{c|}{1.00}      & \multicolumn{1}{c|}{1.00}   & 1.00     & \multicolumn{1}{c|}{1.00}      & \multicolumn{1}{c|}{1.00}   & 1.00     \\
02                                                                       & \multicolumn{1}{c|}{1.00}      & \multicolumn{1}{c|}{0.67}   & 0.80     & \multicolumn{1}{c|}{1.00}      & \multicolumn{1}{c|}{0.33}   & 0.50     & \multicolumn{1}{c|}{1.00}      & \multicolumn{1}{c|}{0.67}   & 0.80     & \multicolumn{1}{c|}{1.00}      & \multicolumn{1}{c|}{1.00}   & 1.00     \\
03                                                                      & \multicolumn{1}{c|}{1.00}      & \multicolumn{1}{c|}{1.00}   & 1.00     & \multicolumn{1}{c|}{1.00}      & \multicolumn{1}{c|}{1.00}   & 1.00     & \multicolumn{1}{c|}{1.00}      & \multicolumn{1}{c|}{0.67}   & 0.80     & \multicolumn{1}{c|}{1.00}      & \multicolumn{1}{c|}{1.00}   & 1.00     \\
04                                                                      & \multicolumn{1}{c|}{1.00}      & \multicolumn{1}{c|}{1.00}   & 1.00     & \multicolumn{1}{c|}{1.00}      & \multicolumn{1}{c|}{0.67}   & 0.80     & \multicolumn{1}{c|}{0.75}      & \multicolumn{1}{c|}{1.00}   & 0.86     & \multicolumn{1}{c|}{1.00}      & \multicolumn{1}{c|}{1.00}   & 1.00     \\
05                                                                       & \multicolumn{1}{c|}{0.75}      & \multicolumn{1}{c|}{1.00}   & 0.86     & \multicolumn{1}{c|}{0.67}      & \multicolumn{1}{c|}{0.67}   & 0.67     & \multicolumn{1}{c|}{1.00}      & \multicolumn{1}{c|}{0.33}   & 0.50     & \multicolumn{1}{c|}{1.00}      & \multicolumn{1}{c|}{0.67}   & 0.80     \\
06                                                                       & \multicolumn{1}{c|}{0.33}      & \multicolumn{1}{c|}{0.33}   & 0.33     & \multicolumn{1}{c|}{0.67}      & \multicolumn{1}{c|}{0.67}   & 0.67     & \multicolumn{1}{c|}{0.67}      & \multicolumn{1}{c|}{0.67}   & 0.67     & \multicolumn{1}{c|}{0.75}      & \multicolumn{1}{c|}{1.00}   & 0.86     \\
07                                                                      & \multicolumn{1}{c|}{1.00}      & \multicolumn{1}{c|}{1.00}   & 1.00     & \multicolumn{1}{c|}{1.00}      & \multicolumn{1}{c|}{1.00}   & 1.00     & \multicolumn{1}{c|}{0.60}      & \multicolumn{1}{c|}{1.00}   & 0.75     & \multicolumn{1}{c|}{1.00}      & \multicolumn{1}{c|}{1.00}   & 1.00     \\
08                                                                       & \multicolumn{1}{c|}{0.75}      & \multicolumn{1}{c|}{1.00}   & 0.86     & \multicolumn{1}{c|}{0.67}      & \multicolumn{1}{c|}{0.67}   & 0.67     & \multicolumn{1}{c|}{0.75}      & \multicolumn{1}{c|}{1.00}   & 0.86     & \multicolumn{1}{c|}{0.75}      & \multicolumn{1}{c|}{1.00}   & 0.86     \\
09                                                                       & \multicolumn{1}{c|}{1.00}      & \multicolumn{1}{c|}{0.67}   & 0.80     & \multicolumn{1}{c|}{0.67}      & \multicolumn{1}{c|}{0.67}   & 0.67     & \multicolumn{1}{c|}{1.00}      & \multicolumn{1}{c|}{0.67}   & 0.80     & \multicolumn{1}{c|}{1.00}      & \multicolumn{1}{c|}{1.00}   & 1.00     \\
10                                                                       & \multicolumn{1}{c|}{0.75}      & \multicolumn{1}{c|}{1.00}   & 0.86     & \multicolumn{1}{c|}{0.75}      & \multicolumn{1}{c|}{1.00}   & 0.86     & \multicolumn{1}{c|}{1.00}      & \multicolumn{1}{c|}{1.00}   & 1.00     & \multicolumn{1}{c|}{1.00}      & \multicolumn{1}{c|}{1.00}   & 1.00     \\
11                                                                      & \multicolumn{1}{c|}{1.00}      & \multicolumn{1}{c|}{1.00}   & 1.00     & \multicolumn{1}{c|}{1.00}      & \multicolumn{1}{c|}{1.00}   & 1.00     & \multicolumn{1}{c|}{1.00}      & \multicolumn{1}{c|}{1.00}   & 1.00     & \multicolumn{1}{c|}{1.00}      & \multicolumn{1}{c|}{1.00}   & 1.00     \\
12                                                                      & \multicolumn{1}{c|}{1.00}      & \multicolumn{1}{c|}{1.00}   & 1.00     & \multicolumn{1}{c|}{1.00}      & \multicolumn{1}{c|}{1.00}   & 1.00     & \multicolumn{1}{c|}{1.00}      & \multicolumn{1}{c|}{0.67}   & 0.80     & \multicolumn{1}{c|}{1.00}      & \multicolumn{1}{c|}{1.00}   & 1.00     \\
13                                                                      & \multicolumn{1}{c|}{1.00}      & \multicolumn{1}{c|}{0.67}   & 0.80     & \multicolumn{1}{c|}{1.00}      & \multicolumn{1}{c|}{1.00}   & 1.00     & \multicolumn{1}{c|}{1.00}      & \multicolumn{1}{c|}{1.00}   & 1.00     & \multicolumn{1}{c|}{1.00}      & \multicolumn{1}{c|}{1.00}   & 1.00     \\
14                                                                      & \multicolumn{1}{c|}{0.75}      & \multicolumn{1}{c|}{1.00}   & 0.86     & \multicolumn{1}{c|}{0.75}      & \multicolumn{1}{c|}{1.00}   & 0.86     & \multicolumn{1}{c|}{0.75}      & \multicolumn{1}{c|}{1.00}   & 0.86     & \multicolumn{1}{c|}{1.00}      & \multicolumn{1}{c|}{1.00}   & 1.00     \\
15                                                                      & \multicolumn{1}{c|}{1.00}      & \multicolumn{1}{c|}{0.67}   & 0.80     & \multicolumn{1}{c|}{1.00}      & \multicolumn{1}{c|}{1.00}   & 1.00     & \multicolumn{1}{c|}{0.50}      & \multicolumn{1}{c|}{0.67}   & 0.57     & \multicolumn{1}{c|}{1.00}      & \multicolumn{1}{c|}{0.67}   & 0.80     \\
16                                                                      & \multicolumn{1}{c|}{1.00}      & \multicolumn{1}{c|}{1.00}   & 1.00     & \multicolumn{1}{c|}{1.00}      & \multicolumn{1}{c|}{1.00}   & 1.00     & \multicolumn{1}{c|}{1.00}      & \multicolumn{1}{c|}{1.00}   & 1.00     & \multicolumn{1}{c|}{1.00}      & \multicolumn{1}{c|}{1.00}   & 1.00     \\ \hline
\end{tabular}
}

\end{table*}

WiAR collected activity data on two T400 laptops using the Intel 5300 NIC. The CSI tool that was developed \cite{halperin2011tool} was employed. One T400 laptop with one antenna transmits outbound signal data, while another with three antennas receives signal data from multiple paths upon reflection. The volunteers were placed in between these two computers to do their activities. The 802.11n NIC cards monitor the channel integrity for each packet that it receives. The CSI is then transmitted to a user-space program for processing by the driver. The Intel 5300 NIC delivers CSI for 30 bands of subcarriers, which are evenly divided across the 56 subcarriers of a 20 MHz channel or the 114 carriers of a 40 MHz channel \cite{halperin2011tool}.



\subsection{Different Factors}

The frequencies used among commercial Wi-Fi systems range from 2.4GHz to 5GHz. WiAR used 20MHz bandwidth with 30 subcarriers at 5GHz, which provides higher stability than 2.4GHz. WiAR obtained CSI data for a single transmitter antenna and three receiver antennas. The use of multiple antennas extends the range of the Wi-Fi data communication system, but it still introduces complications in the detection process. Despite the additional challenges, data from multiple antennas gives more information than just a single antenna transmitter-receiver system, enhancing the recognition system.

Ten volunteers carry out all activities for added robustness. Because different individuals need varying amounts of time to complete their activities, having a large dataset of volunteers is essential. Also, the body's shape, height, and pattern of activity are all essential factors in the recognition procedure. WiAR collected 30 samples from each of the volunteers. As indicated \cite{guo2019wiar}, the effect of volunteers on CSI data is examined in terms of sex, height, weight, and experience. 

WiAR created three distance levels to investigate the effects of different distances between the transmitter and the receiver on human activity recognition. The distance factor data is collected for three distances; the three distances are 1 m, 3 m, and 6 m. The hall served as the experimental setting, and the environment was more challenging owing to the combination of glass walls and elevators. WiAR also created three levels for the height between the receiver and the floor to assess the influence of varying heights on human activity recognition. The measurements are 60 cm, 90 cm, and 120 cm, respectively.

\subsection{Performance Analysis}
\begin{figure*}[!ht]
         \centering
         \begin{subfigure}[b]{0.33\textwidth}
         \includegraphics[width = \textwidth]{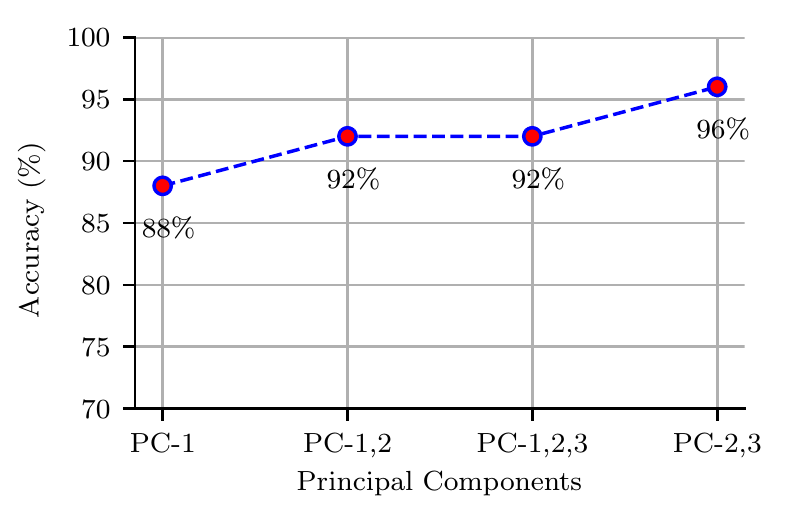}
         \caption{}
         \label{pca_comp1}
         \end{subfigure}
         \centering
         \begin{subfigure}[b]{0.33\textwidth}
         \includegraphics[width = \textwidth]{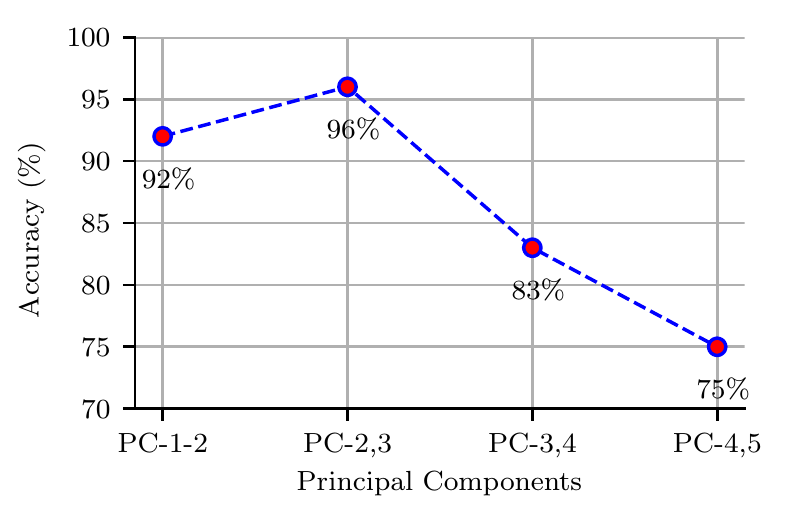}
         \caption{}
         \label{pca_comp2}
         \end{subfigure}
         \centering
         \begin{subfigure}[b]{0.33\textwidth}
         \includegraphics[width = \textwidth]{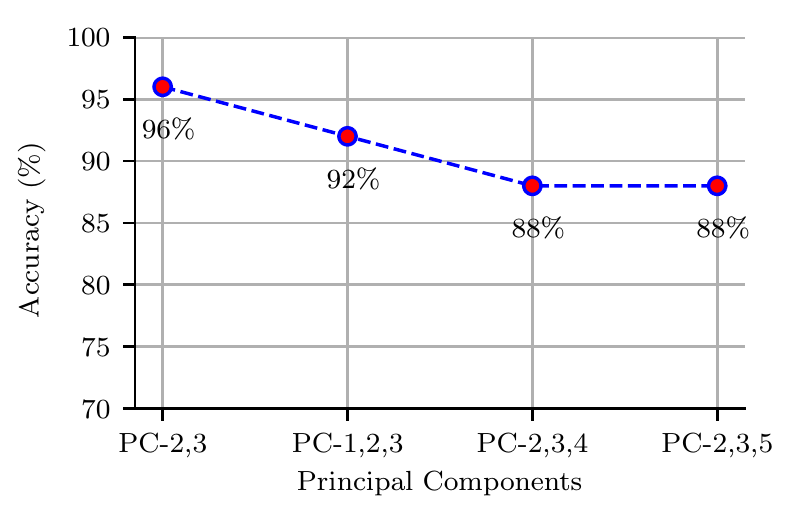}
         \caption{}
         \label{pca_comp3}
         \end{subfigure}

\caption{Performance comparison of different combinations of principal components (PCs) using DWT feature extraction and 1D-CNN for one individual}
\label{pca_comp}
\end{figure*}

\begin{table}[!ht]
\caption{Overall Performance of PCWCNN approach}
\label{wcnn_table}
\centering
\begin{tabular}{|l|c|c|c|}
\hline
\multicolumn{1}{|c|}{Activity Class} & Precision & Recall & F1-Score \\ \hline
01. Horizontal Arm Wave              & 1.00      & 1.00   & 1.00     \\
02. High Arm Wave                    & 1.00      & 0.89   & 0.94     \\
03. Two Hands Wave                   & 0.95      & 1.00   & 0.97     \\
04. High Throw                       & 1.00      & 0.89   & 0.94     \\
05. Draw X                           & 0.90      & 1.00   & 0.95     \\
06. Draw Tick                        & 1.00      & 1.00   & 1.00     \\
07. Toss Paper                       & 0.82      & 1.00   & 0.90     \\
08. Forward Kick                     & 0.85      & 0.94   & 0.89     \\
09. Side Kick                        & 1.00      & 0.94   & 0.97     \\
10. Bend                             & 1.00      & 1.00   & 1.00     \\
11. Hand Clap                        & 1.00      & 0.94   & 0.97     \\
12. Walk                             & 1.00      & 0.94   & 0.97     \\
13. Phone Call                       & 1.00      & 0.89   & 0.94     \\
14. Drink Water                      & 0.94      & 0.89   & 0.91     \\
15. Sit Down                         & 0.90      & 1.00   & 0.95     \\
16. Squat                            & 1.00      & 0.94   & 0.97     \\ \hline
Macro Average                        & 0.96      & 0.95   & 0.96     \\ \hline
\end{tabular}

\end{table}

\begin{table}[!ht]
\centering
\caption{Recognition accuracy comparison of different approaches}
\label{wcnn_acc}
\begin{tabular}{|c|c|c|c|}
\hline
\textbf{\begin{tabular}[c]{@{}c@{}}Recognition \\ Approach\end{tabular}} & \textbf{\begin{tabular}[c]{@{}c@{}}Number of \\ Volunteers\end{tabular}} & \textbf{\begin{tabular}[c]{@{}c@{}}Number of \\ Activities\end{tabular}} & \textbf{\begin{tabular}[c]{@{}c@{}}Recognition\\  Accuracy\end{tabular}} \\ \hline
\textbf{CARM}                                                            & 25                                                                       & 8                                                                        & 96.00\%                                                                  \\ \hline
\textbf{LCED}                                                            & 10                                                                       & 16                                                                       & 95.00\%                                                                  \\ \hline
\textbf{WiFall}                                                          & 10                                                                       & 4                                                                        & 94.00\%                                                                  \\ \hline
\textbf{PCWCNN}                                                            & 10                                                                       & 16                                                                       & 95.50\%                                                                  \\ \hline
\end{tabular}
\end{table}

\begin{figure}[!ht]
         \centering
         \includegraphics[width = 0.45\textwidth]{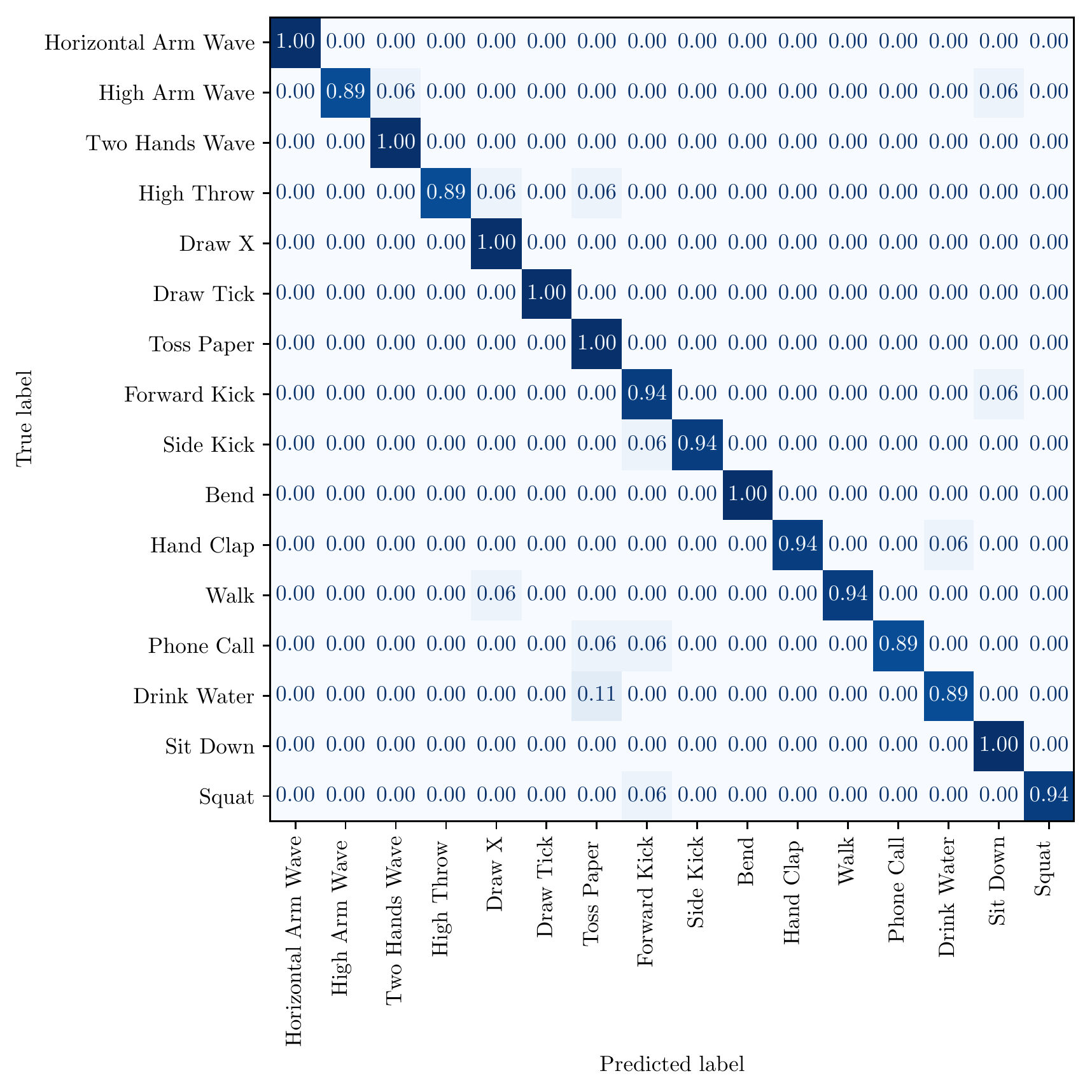}
         \caption{Confusion matrix generated from PCWCNN approach}
\label{wcnn_con_mat}
\end{figure}

\begin{figure*}[!ht]
         \centering
         \begin{subfigure}[b]{0.46\textwidth}
         \includegraphics[width=\textwidth]{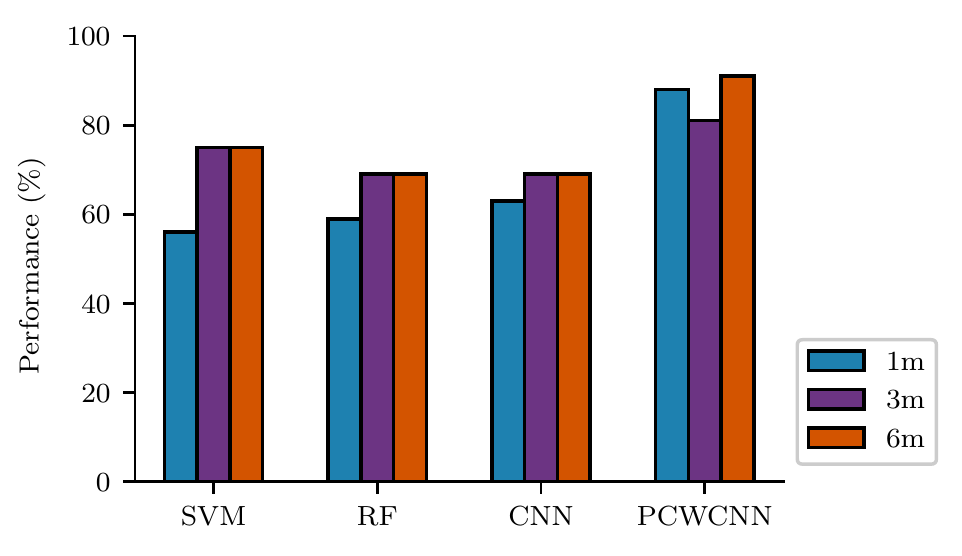}
         \caption{Effect of distance}
         \end{subfigure}
         \begin{subfigure}[b]{0.46\textwidth}
         \includegraphics[width=\textwidth]{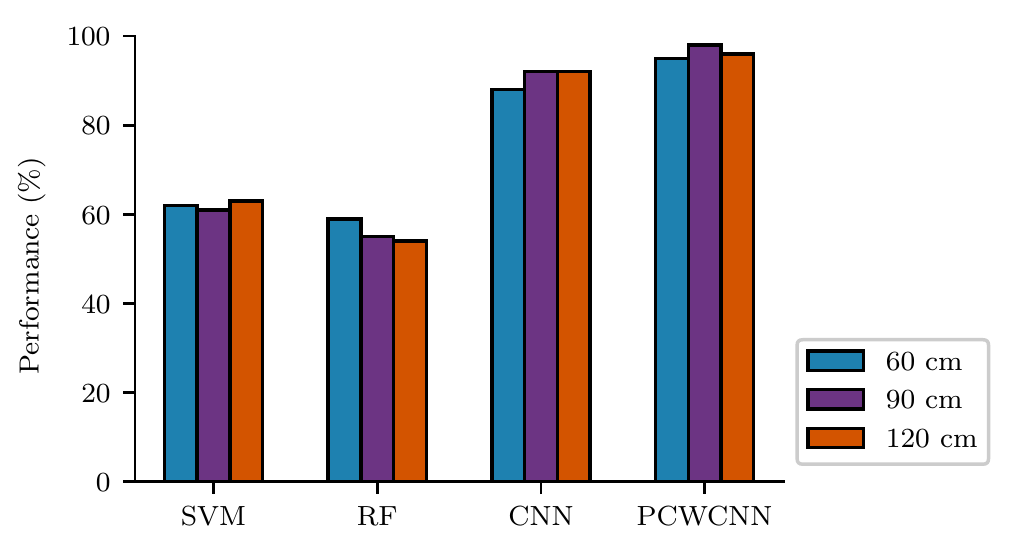}
         \caption{Effect of height}
         \end{subfigure}

\caption{Effect of distance and height on PCWCNN}
\label{wcnn_distance_height}
\end{figure*}
A few reference models (SVM, RF, CNN) are compared to the proposed PCWCNN framework. Support Vector Machine (SVM) is a supervised nonparametric algorithm that performs effectively with high-dimensional feature spaces. The Random Forest algorithm combines several randomized decision trees and aggregates their predictions to produce a robust learner. For RF, the classifier's number of trees was set to 200, and the maximum depth was set to 20. Convolution Neural Network (CNN) is a commonly deployed DL technique that includes shift-invariance, which is crucial since the activity segment might occur at any moment. One of the difficulties in employing CNNs is that the CSI signal recorded from the experiments may have a variable duration. Zero padding is one method of getting around the CNN input layer's limitation. This study uses a One-dimensional Convolutional Neural Network (1D-CNN) model. The model involves a 384-element vector as input and outputs a vector containing probabilities for each potential activity type. Before training the models, 20\% of the data was reserved as the test set, and the rest was used for training.

Fig. \ref{pca_dwt_fig} shows a performance comparison of the reference models for two cases: using all the  subcarriers and using subcarrier fusion. The performance disparity between the two methods is negligible. It can be argued that the PCA-based subcarrier fusion approach is more appropriate in terms of efficiency with little to no performance sacrifice given the subcarrier fusion method's strength in dimensionality reduction. Table \ref{pca_dwt} compares the accuracy, recall, and F1-score of different approaches for one volunteer using subcarrier fusion. The accuracy of the PCA-based subcarrier fusion method using CNN is 96\%, compared to 94\% for the all subcarriers method. SVM and RF perform similarly in both cases, however KNN performance degrades when using the subcarrier fusion method.

The performance comparison of several principal component groupings is demonstrated in Fig. \ref{pca_comp}. We applied a 1D-CNN network together with DWT-based feature extraction to illustrate this comparison. The effect of combining the first principal component (PC1) with other components is depicted in Fig. \ref{pca_comp1}. It is obvious that the first principal component is insufficient. The accuracy improves from 88\% to 92\% when PC2 or PC3 are integrated with PC1. However, PC2 and PC3 pair performs significantly better than any other combination that includes PC1. Comparisons between several PC pairings are shown in Fig. \ref{pca_comp2}. Evidently, PC-2,3 pair outperforms others. PC-4,5 pair performs much worse (75\%), which points to noise in the later-end PCs. Fig. \ref{pca_comp3} compares the performance of combining various PCs with PC-2,3. Performance always degrades regardless of which PC is added to the pair PC-2,3. The rationale for using the PC-2,3 combination in our proposed PCWCNN model is demonstrated by this comparison study.

The accuracy, recall, and F1-score of all the volunteers using the proposed PCWCNN technique are shown in Table \ref{wcnn_table}. The results show the superiority of this method over previous reference models. Fig. \ref{wcnn_con_mat} depicts the approach's confusion matrix. Whereas the majority of activities were competently recognized, the detection accuracy of some of them was low (89\%), which might be attributed to similarities between them or related activities. High arm waves, high throw, phone call, and drink water are all upper body activities. Although these activities can be identified visually, using CSI alone would make it substantially more challenging. Additionally, the activity patterns of these activities are very dependent on the specific individual. Therefore, in comparison, the detection accuracy of these activities is somewhat lower. Moreover, compared to the results obtained for one volunteer, the overall performance findings demonstrate a considerable performance decline. The reasoning can be deduced due to some individuals' poor quality data and the system becoming more complex due to the inclusion of more volunteers.

Table \ref{wcnn_acc} compares the accuracy of the methodologies described in previous literature, as well as a synopsis of the dataset that was used. In terms of accuracy, the PCWCNN method yields promising results, with accuracy surpassing 95\%. Because additional classes of activities complicate the problem, the best analogy is the LCED technique, which has a 95\% overall accuracy.

\subsection{Effect of Distance and Height for PCWCNN}
The performance of the proposed PCWCNN technique for various distance and height data is shown in Fig. \ref{wcnn_distance_height}. The distance factor data is collected for three distances: 1 m, 3 m, and 6 m. The results outperform reference models considerably. Performance for 6m data is the best for distance factor data. But Wi-Fi signals have a transmission range of fewer than 15 meters and a detecting range of fewer than 5 meters \cite{guo2021towards}. Therefore the results obtained are contradictory to the general idea.

Since different height data correlates to various body sections, the WiAR dataset provides various height data. The lower body corresponds to 60 cm in height. The whole body corresponds to 90 cm. Upper-body activities correlate to a height of 120 cm. Regarding height factor data, the highest performance comes from 90 cm data, which can be linked to the fact that at this height, all activities relating to all heights are given equal attention.

         

\section{Conclusion}
In this paper, we proposed a novel human activity recognition method that is accurate and robust for real-time applications. We achieved this by employing an adaptive activity segmentation algorithm, PCA- and DWT-based preprocessing and CNN-based classification. We performed extensive experimentation on a real dataset, and empirically compared the performance of our method against several recent approaches. We demonstrated that our method comfortably outperforms the existing ones. An interesting future work could be enhancing the resilience of our proposed method against environmental changes. Another direction could be incorporating formal privacy guarantees in our method, as CSI data correlated with human movement is potentially privacy-sensitive.



\bibliographystyle{unsrtnat}
\bibliography{references}





\end{document}